%% file: Main.tex
\documentclass[sigconf,natbib=true]{acmart}

\copyrightyear{2026}
\acmYear{2026}
\setcopyright{cc}
\setcctype{by}
\acmConference[SIGIR '26]{Proceedings of the 49th International ACM SIGIR Conference on Research and Development in Information Retrieval}{July 20--24, 2026}{Melbourne, VIC, Australia}
\acmBooktitle{Proceedings of the 49th International ACM SIGIR Conference on Research and Development in Information Retrieval (SIGIR '26), July 20--24, 2026, Melbourne, VIC, Australia}
\acmDOI{10.1145/3805712.3809958}
\acmISBN{979-8-4007-2599-9/2026/07}

\usepackage{libertine}
\usepackage{booktabs}
\usepackage{multirow}
\usepackage{pifont}
\usepackage{xspace}
\usepackage{enumitem}
\usepackage{balance}
\usepackage[table]{xcolor} 
\usepackage[most]{tcolorbox}
\usepackage[normalem]{ulem}
\useunder{\uline}{\ul}{}

\begin{document}

\title{Pretraining Exposure Explains Popularity Judgments in Large Language Models}


\author{Jamshid Mozafari}
\authornote{Corresponding Author.}
\orcid{0000-0003-4850-9239}
\affiliation{%
  \institution{University of Innsbruck}
 \city{Innsbruck}
  \country{Austria}
  }
\email{jamshid.mozafari@uibk.ac.at}

\author{Bhawna Piryani}
\orcid{0009-0005-3578-2393}
\affiliation{%
  \institution{University of Innsbruck}
  \city{Innsbruck}
  \country{Austria}
  }
\email{bhawna.piryani@uibk.ac.at}

\author{Adam Jatowt}
\orcid{0000-0001-7235-0665}
\affiliation{%
  \institution{University of Innsbruck}
  \city{Innsbruck}
  \country{Austria}
  }
\email{adam.jatowt@uibk.ac.at}



\begin{abstract}
Large language models (LLMs) exhibit systematic preferences for well-known entities, a phenomenon often attributed to popularity bias. However, the extent to which these preferences reflect real-world popularity versus statistical exposure during pretraining remains unclear, largely due to the inaccessibility of most training corpora. We provide the first direct, large-scale analysis of popularity bias grounded in fully observable pretraining data. Leveraging the open OLMo models and their complete pretraining corpus, Dolma, we compute precise entity-level exposure statistics across 7.4 trillion tokens. We analyze 2,000 entities spanning five types (Person, Location, Organization, Art, Product) and compare pretraining exposure against Wikipedia pageviews and two elicited LLM popularity signals: direct scalar estimation and pairwise comparison.
Our results show that pretraining exposure strongly correlates with Wikipedia popularity, validating exposure as a meaningful proxy for real-world salience during the training period. More importantly, we find that LLM popularity judgments align more closely with exposure than with Wikipedia, especially when elicited via pairwise comparisons. This alignment is strongest for larger models and persists in the long tail, where Wikipedia popularity becomes unreliable. Overall, our findings demonstrate that popularity priors in LLMs are primarily shaped by pretraining statistics rather than external popularity signals, offering concrete evidence that data exposure plays a central role in driving popularity bias.
\end{abstract}





\maketitle

\section{Introduction}\label{s:introduction} 
\input{figures/pipeline}

Large language models (LLMs)~\cite{grattafiori2024llama, yang2025qwen3, team2024gemma} exhibit systematic preferences for well-known entities, often treating them as more salient or important than less prominent ones~\cite{lehmann2025knowing, wang-etal-2023-causal}. While such popularity priors are widely observed, their origin remains unclear~\cite{wang-etal-2023-causal, 10.1145/3597307}. In particular, it is not well understood whether these priors reflect real-world popularity signals or instead emerge from statistical properties of the data used during pretraining.

A natural hypothesis is that LLM popularity judgments are driven by \textbf{pretraining exposure}---that is, the frequency and distribution with which entities and their aliases appear in the training corpus\footnote{In this work, we use \textit{pretraining exposure} specifically to denote frequency-based occurrence in the training data, distinct from prior uses of exposure in membership inference or train--test mismatch settings.}. Testing this hypothesis is challenging because the pretraining data of released LLMs are typically not publicly accessible, and the sheer scale of these corpora further precludes exhaustive indexing and direct measurement of entity exposure~\cite{shi2023detecting, zhang-etal-2024-pretraining, balloccu-etal-2024-leak}. Consequently, prior analyses~\cite{prato-etal-2024-large, dodge-etal-2021-documenting, lehmann2025knowing, wang-etal-2023-causal, kadavath2022language} have relied on indirect evidence or downstream model behavior, making it difficult to disentangle exposure effects from other confounding factors.

Prior work has examined popularity bias~\cite{shuster-etal-2022-language}, memorization~\cite{carlini2021extracting}, and exposure effects~\cite{10.1145/3357713.3384290} in LLMs, showing that model outputs often reflect frequency patterns in pretraining data rather than purely semantic or factual considerations~\cite{dodge-etal-2021-documenting, kadavath2022language, lehmann2025knowing, wang-etal-2023-causal}. Related studies further demonstrate that data scale and distribution shape model behavior, leading to systematic preferences for frequently observed entities or concepts~\cite{prato-etal-2024-large}. More recently, these observations have been formalized under the notion of \textit{pretraining data exposure}, which unifies memorization, data contamination, and membership inference under a common lens of what models have ``seen'' during training~\cite{tong2025pde_survey}. At the same time, exposure is not only a property of the data but also of the training process: exposure bias arising from discrepancies between training and inference can amplify pretraining-driven patterns, particularly in distillation settings~\cite{pozzi2025distillation_exposure_bias}. Similarly, in model updating scenarios, LLMs tend to favor pretraining knowledge even after fine-tuning, unless explicitly mitigated~\cite{yu2023self_information_update}. Together, these findings suggest that both data exposure and procedural biases reinforce the influence of pretraining statistics.

\input{figures/prompt_aliases}

In this work, we directly study the relationship between pretraining exposure, real-world popularity, and LLM popularity judgments using the fully open pretraining data of OLMo, known as Dolma~\cite{soldaini-etal-2024-dolma}. OLMo~\cite{olmo2025olmo} is a fully open LLM for which the complete pretraining corpora are publicly available, enabling direct analysis of model behavior with respect to its training data. Figure~\ref{fig:pipeline} illustrates our analysis pipeline. We first index the OLMo pretraining corpora using the Infini-Gram toolkit~\cite{liuinfini} to compute entity-level exposure statistics. Entities are sampled from Wikidata\footnote{\url{https://www.wikidata.org}} across multiple types (Person, Location, Organization, Art, and Product), with validated aliases used to ensure accurate frequency estimation. We then compare exposure scores against Wikipedia pageview statistics as an external measure of real-world popularity. Finally, we probe OLMo-3-7B and OLMo-3.1-32B using two elicitation strategies---direct popularity estimation (\textit{Directly}) and pairwise comparison (\textit{Comparison})---to analyze how models express popularity judgments.

Our work provides the first direct, large-scale measurement of entity-level pretraining exposure and systematically compares it against both real-world popularity signals and LLM-generated popularity judgments. Our results show that pretraining exposure strongly correlates with Wikipedia popularity across entity types, validating exposure as a meaningful popularity proxy. Moreover, LLM popularity judgments---especially when elicited via \textit{Comparison}---align more closely with exposure than with Wikipedia popularity, with notable differences between high- and low-popularity entities across entity types. These findings indicate that LLM popularity priors are largely explained by pretraining statistics in this setting, particularly when real-world popularity signals are weak or noisy. To facilitate transparency and enable further research, we publicly release our code, processed data, and full analysis pipeline—including entity sets, exposure statistics, and evaluation scripts—on a GitHub repository\footnote{\url{https://github.com/DataScienceUIBK/Pretraining-Exposure-Popularity}}.

\input{figures/prompt_directly}

\section{Dataset}\label{s:dataset}
To enable a reliable analysis, a set of entities with corresponding exposure scores derived from OLMo pretraining data is required. To this end, the complete OLMo training corpus—including \textit{Pretraining}\footnote{\url{https://huggingface.co/datasets/allenai/dolma3_mix-6T}}, \textit{Midtraining}\footnote{\url{https://huggingface.co/datasets/allenai/dolma3_dolmino_mix-100B-1125}}, and \textit{Long Context}\footnote{\url{https://huggingface.co/datasets/allenai/dolma3_longmino_mix-100B-1125}} datasets—is indexed using the Infini-Gram toolkit~\cite{liuinfini}. The combined corpus comprises approximately 27 terabytes of data containing 7.4 trillion tokens. After indexing, the corpus is partitioned into 128 shards with a total indexed size of approximately 50 terabytes. This corpus has been originally used to train both OLMo-3-7B and OLMo-3.1-32B, enabling a controlled comparison across model scales.

Entities are extracted using Wikidata randomly. The most recent Wikidata dump available as of 24/01/2026 is downloaded\footnote{\url{https://dumps.wikimedia.org/wikidatawiki/entities}}, from which 5{,}000 entities and their associated aliases are sampled for each entity type: Person, Location, Organization, Art, and Product. Since Wikidata aliases may be ambiguous or noisy—for example, the alias \textit{us} for the entity \textit{United States of America}—all aliases are validated and filtered using GPT-4.1~\cite{achiam2023gpt} with the prompt shown in Figure~\ref{fig:prompt_aliases}. Figure~\ref{fig:long_tail} shows that the resulting entity distributions exhibit a long-tail pattern, consistent with the natural popularity distribution of entities~\cite{mallen-etal-2023-trust, petroni-etal-2019-language}.

\input{figures/long_tail}

Given the indexed corpus and validated entity aliases, an exposure score is computed for each entity by counting the frequency of the entity and its aliases in the indexed data. The Infini-Gram toolkit handles overlap between an entity and its aliases and prevents double counting\footnote{Infini-Gram supports AND/OR operators in conjunctive normal form (CNF).}. For each entity type, entities are sorted by exposure score in ascending order, and 400 entities are selected\footnote{We choose 400 because the pairwise \textit{Comparison} method would require about 750 million prompts for 5{,}000 entities, which is infeasible; using 400 reduces this to about 4.8 million.}: the 200 lowest-exposure entities are designated as \textit{sparse entities}, while the 200 highest-exposure entities are designated as \textit{popular entities}. Across all five entity types, this results in a total of 2{,}000 entities.

\input{figures/pairwise_accuracy}

\section{Popularity Signals}\label{s:popularity_signals}

To analyze popularity signals, three additional signals are considered alongside pretraining exposure: \textit{Wikipedia}, \textit{Directly}, and \textit{Comparison}. Each signal is described below.

\paragraph{Wikipedia}
This popularity signal is computed using the number of pageviews of the Wikipedia page for each entity using the Wikipedia Pageviews API\footnote{\url{https://pageviews.wmcloud.org}}. Pageview counts are aggregated from July~1,~2015—the earliest date for which pageview statistics are available—through December~31,~2024, which corresponds to the cutoff date of the OLMo pretraining corpus. This temporal restriction is necessary because the popularity of some entities may increase substantially after the pretraining period (e.g., \textit{Greenland}\footnote{\url{https://en.wikipedia.org/wiki/Greenland}}), and such changes would not be reflected in the training data.

\input{figures/prompt_comparison}
\input{tables/correlation}

\paragraph{Directly}
To compute this popularity signal, OLMo is prompted using the template shown in Figure~\ref{fig:prompt_directly}, which asks the model to estimate the popularity of an entity and its aliases using a numerical score between 0 and 1000. Due to the non-deterministic nature of LLM generation, LLM is queried three times per entity, and the final \textit{Directly} score is obtained by averaging the generated scores. This repetition-based aggregation yields more stable estimates~\cite{wangself}.

\paragraph{Comparison}
To compute this popularity signal, the OLMo model is prompted using the template shown in Figure~\ref{fig:prompt_comparison} to perform pairwise popularity comparisons between two entities and their aliases. The model predicts which of the two entities is more popular. For each entity pair, LLM is queried three times, and the final comparison outcome is determined by majority vote. The prompt additionally requests a brief justification for each decision, as prior work has shown that generating explanations can lead to more reliable model outputs~\cite{huang-etal-2023-large}. The resulting pairwise preferences are subsequently converted into listwise popularity scores using the Bradley--Terry model~\cite{bradley-terry}.

\section{Experimental Setup}\label{s:experimental_setup}
Indexing the OLMo corpora is performed on 16 compute nodes, each equipped with 380 CPU cores and 760~GB of RAM, over a period of seven days. Approximately 100~TB of storage is used to store the raw corpora, indexed shards, Wikidata dumps, and auxiliary files. Model inference experiments are conducted on four GPU nodes, each containing four NVIDIA H100 GPUs with 94~GB of memory. All experiments are implemented using PyTorch~\cite{pytorch} and the Transformers library~\cite{wolf-etal-2020-transformers}. Two OLMo model variants are evaluated: OLMo-3-7B\footnote{\url{https://huggingface.co/allenai/Olmo-3-7B-Instruct}} and OLMo-3.1-32B\footnote{\url{https://huggingface.co/allenai/Olmo-3.1-32B-Instruct}}, both trained on the indexed corpora.  Other LLMs are not considered, as their pretraining data are usually not publicly available, preventing direct measurement of pretraining exposure.

\section{Results and Analysis}\label{s:results_analysis}

Table~\ref{tbl:correlation} reports Spearman’s rank correlations~\cite{spearman-correlation} between each popularity signal and \textit{pretraining exposure}, across entity types and two OLMo model scales (7B and 32B). Since \textit{Wikipedia} is an external signal computed independently of model inference, its correlation with exposure is identical across model sizes. In contrast, the \textit{Directly} and \textit{Comparison} signals reflect model behavior and therefore vary with scale.

\paragraph{\textbf{Exposure is a strong proxy for Wikipedia popularity}}
Across \textit{all entities}, pretraining exposure exhibits a strong positive correlation with Wikipedia pageviews for every entity type, with Spearman’s $\rho$ ranging from 0.688 (Product) to 0.826 (Location), and an overall average of 0.7558. This result validates exposure as a meaningful proxy for real-world popularity during the pretraining period, and indicates that the indexed corpus captures broad real-world salience patterns despite being composed of heterogeneous web-scale sources.

\paragraph{\textbf{LLM popularity judgments align more with exposure than with Wikipedia}}
When considering \textit{all entities}, LLM-generated popularity signals consistently align more closely with exposure than with Wikipedia pageviews. The \textit{Directly} signal shows moderate exposure alignment for OLMo-3-7B (average $\rho{=}0.6144$) and improves substantially for OLMo-3.1-32B (average $\rho{=}0.7534$), approaching the exposure--Wikipedia correlation.

The \textit{Comparison} signal yields the strongest alignment with exposure at both model scales, achieving average correlations of 0.7372 for 7B and 0.795 for 32B. For OLMo-3-7B, pairwise elicitation closes most of the gap to Wikipedia (0.7372 vs.\ 0.7558) and substantially outperforms direct scalar estimation. For OLMo-3.1-32B, \textit{Comparison} becomes the best-performing signal overall, exceeding both \textit{Directly} and \textit{Wikipedia} in exposure alignment. These results indicate that LLM popularity priors are more faithfully revealed through pairwise comparison than through direct numerical scoring.

\paragraph{\textbf{Sparse entities: Wikipedia appears to become unreliable, exposure alignment persists}}
A clear regime change emerges for \textit{sparse entities} (the 200 lowest-exposure entities per type). In this regime, Wikipedia pageviews show near-zero and occasionally negative correlations with exposure (average $\rho{=}0.1012$ across entity types), suggesting that aggregated pageview statistics provide a questionable popularity signal for long-tail entities.

In contrast, LLM-based popularity signals retain positive alignment with exposure. For OLMo-3-7B, \textit{Directly} reaches an average $\rho{=}0.2356$, while \textit{Comparison} further improves alignment to 0.2836. The same pattern holds for OLMo-3.1-32B, with average correlations of 0.243 and 0.3088 for \textit{Directly} and \textit{Comparison}, respectively. The persistence of exposure alignment when Wikipedia fails suggests that model priors for rare entities are primarily governed by pretraining frequency rather than external popularity signals.

\paragraph{\textbf{Popular entities: Wikipedia improves, but comparison remains strongest}}
For \textit{popular entities} (the 200 highest-exposure entities per type), Wikipedia pageviews become more informative, with an average correlation of 0.4082. However, LLM-based signals—particularly \textit{Comparison}—continue to show stronger alignment with exposure. Pairwise elicitation achieves average correlations of 0.4508 for OLMo-3-7B and 0.5096 for OLMo-3.1-32B, consistently outperforming \textit{Directly} (0.3028 and 0.3584, respectively).

These results indicate that even when entities are broadly well-known, pairwise comparison better captures consistent ranking preferences, yielding exposure-consistent popularity judgments beyond what is obtained through direct scoring.


\paragraph{\textbf{Effect of model scale}}
Scaling from OLMo-3-7B to OLMo-3.1-32B systematically increases alignment with pretraining exposure for both elicitation methods. The improvement is most pronounced for \textit{Directly} on \textit{all entities}, where the average Spearman correlation increases from 0.6144 to 0.7534 (gain of 0.1390), indicating more stable and exposure-consistent scalar estimates. For \textit{Comparison}, the average correlation increases from 0.7372 to 0.795 on all entities, from 0.4508 to 0.5096 on popular entities, and from 0.2836 to 0.3088 on sparse entities. These consistent gains suggest that increased model capacity sharpens exposure-based popularity priors rather than fundamentally changing their source.

\paragraph{\textbf{Pairwise reliability and listwise scoring}}
Figure~\ref{fig:pairwise_accuracy} summarizes the reliability of the pairwise judgments used by the \textit{Comparison} signal. 
This reliability helps explain why \textit{Comparison} consistently yields the strongest alignment with pretraining exposure across entity types, regimes, and model scales.

\section{Conclusion}\label{s:conclusion}

We showed that popularity judgments in large language models are largely explained by \textit{pretraining exposure}. Using the fully open OLMo corpus, we found that exposure strongly correlates with Wikipedia popularity, and that LLM-generated popularity judgments—especially under pairwise elicitation—align more closely with exposure than with Wikipedia signals. This effect is strongest for larger models and persists in the long tail, where external popularity measures become unreliable. 

Our results suggest that popularity bias in LLMs can be traced to measurable properties of their training data. As future work, we plan to extend this analysis beyond entities to broader notions of popularity, including events, topics, claims, and contextualized content, in order to better understand how exposure shapes model behavior at different levels of abstraction. Additionally, investigating methods to mitigate exposure-driven biases could help improve fairness and diversity in model outputs. 

\begin{acks}
The computational results presented here have been achieved, in part, using the LEO HPC infrastructure of the University of Innsbruck and the Austrian Scientific Computing (ASC) infrastructure.
\end{acks}

\bibliographystyle{ACM-Reference-Format}
\balance
\bibliography{Main}

\end{document}

%% file: figures/pipeline.tex
\begin{figure}[]
	\centering
	\includegraphics[width=0.8\columnwidth]{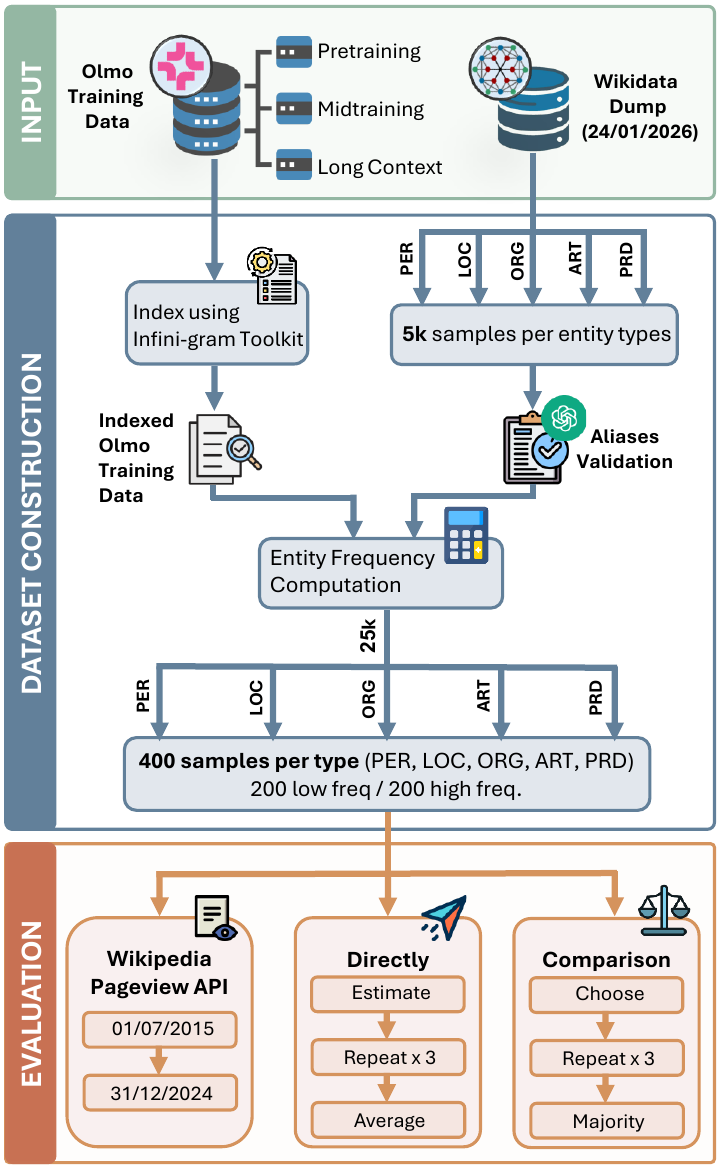}
    \caption{Illustration of the full analysis workflow. The OLMo pretraining corpus is indexed to derive entity exposure statistics, followed by entity sampling and alias validation from Wikidata. Popularity signals are then evaluated using Wikipedia pageviews and two LLM-based methods: direct scalar estimation and pairwise comparison.}
	\label{fig:pipeline}
\end{figure}

%% file: figures/prompt_aliases.tex
\begin{figure}[t]
\centering
\begin{minipage}{\columnwidth}
\begin{tcolorbox}[
  colback=gray!5,
  colframe=black!70,
  boxrule=0.4pt,
  arc=2pt,
  left=6pt,
  right=6pt,
  top=6pt,
  bottom=6pt,
  fonttitle=\bfseries,
  enhanced
]
\small
You are given a target entity.
For each option, decide whether it is an alias that refers exclusively to the same entity and does not commonly refer to any other distinct entities or concepts.

Target entity: $<$\texttt{Entity}$>$

Options: $<$\texttt{Options}$>$

Output format requirement:

Respond with only a valid JSON array of integers.
Do not include any explanations, text, markdown, or formatting.
\end{tcolorbox}
\end{minipage}
\caption{Alias validation prompt. The model is provided with a target entity (\texttt{<Entity>}) and a set of candidate aliases (\texttt{<Options>}), and must identify those that exclusively refer to the same entity without ambiguity.}
\label{fig:prompt_aliases}
\end{figure}

%% file: figures/prompt_directly.tex
\begin{figure}[t]
\centering
\begin{minipage}{\columnwidth}
\begin{tcolorbox}[
  colback=gray!5,
  colframe=black!70,
  boxrule=0.4pt,
  arc=2pt,
  left=6pt,
  right=6pt,
  top=6pt,
  bottom=6pt,
  fonttitle=\bfseries,
  enhanced
]
\small
You are a popularity estimator based on your data and general knowledge. Estimate the popularity of the entity below.

Return only a single integer between 0 and 1000, with no explanation.

Entity: $<$\texttt{Entity}$>$

Score (0 to 1000): 
\end{tcolorbox}
\end{minipage}
\caption{Direct popularity estimation prompt. The model receives an entity (\texttt{<Entity>}) and outputs a scalar score in the range [0, 1000] representing its estimated popularity.}
\label{fig:prompt_directly}
\end{figure}

%% file: figures/long_tail.tex
\begin{figure}[]
	\centering
	\includegraphics[width=0.8\columnwidth]{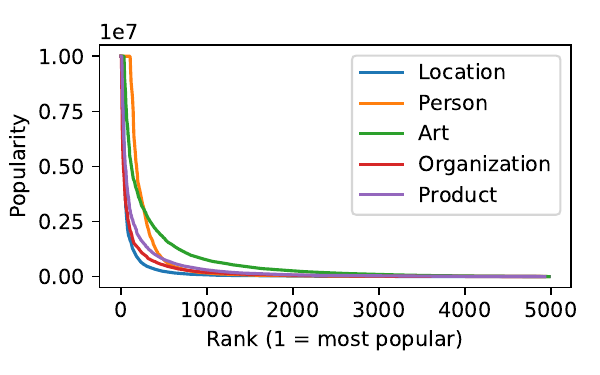}
    \caption{Long-tail distribution of entity frequency (pretraining exposure) across entity types, computed from occurrences in the indexed training corpus and used as a proxy for popularity in our analysis.}
	\label{fig:long_tail}
\end{figure}

%% file: figures/pairwise_accuracy.tex
\begin{figure}[]
	\centering
	\includegraphics[width=\columnwidth]{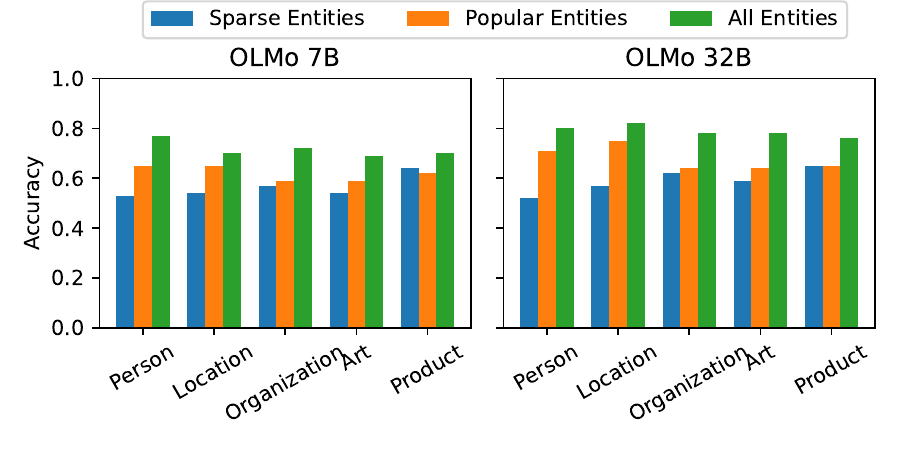}
    \caption{Pairwise popularity comparison accuracy across entity types and popularity groups.}
	\label{fig:pairwise_accuracy}
\end{figure}

%% file: figures/prompt_comparison.tex
\begin{figure}[t]
\centering
\begin{minipage}{\columnwidth}
\begin{tcolorbox}[
  colback=gray!5,
  colframe=black!70,
  boxrule=0.4pt,
  arc=2pt,
  left=6pt,
  right=6pt,
  top=6pt,
  bottom=6pt,
  fonttitle=\bfseries,
  enhanced
]
\small
You are a popularity estimator with access to general world knowledge. Given the two entities below, determine which one is more popular.

Select the correct option and briefly justify your choice.

Return your answer ONLY in valid JSON format, strictly following the template below.

Options:

\textbf{1}: \texttt{Entity$_1$} is more popular than \texttt{Entity$_2$}

\textbf{2}: \texttt{Entity$_2$} is more popular than \texttt{Entity$_1$}

Output:
\begin{verbatim}
{
  "e1": "Entity_1",
  "e2": "Entity_2",
  "justification": "Short explaination of your decision.",
  "option": 1 or 2
}
\end{verbatim}

\end{tcolorbox}
\end{minipage}
\caption{Prompt used for the \textit{Comparison} method. \texttt{<Entity\_1>} and \texttt{<Entity\_2>} denote the two entities being compared. \texttt{<Justification>} provides a brief explanation generated by the LLM, and \texttt{<option>} indicates the entity predicted to be more popular.
}
\label{fig:prompt_comparison}
\end{figure}

%% file: tables/correlation.tex
\newcommand{\myhigh}{\cellcolor{gray!30}} 

\begin{table*}[t]
\centering
\caption{Spearman correlation ($\rho$) between popularity signals and entity exposure across entity types and model scales. Results are reported for OLMo-3-7B and OLMo-3.1-32B on all, sparse, and popular entities. The \textit{Directly} and \textit{Comparison} signals correspond to absolute and pairwise LLM popularity judgments, respectively. Higher values indicate stronger alignment with pretraining exposure. \textcolor{gray}{Gray} cells highlight the maximum value, and underlined values denote the highest average correlation.}

\label{tbl:correlation}

\resizebox{\textwidth}{!}{
\begin{tabular}{@{}lcccccccccccc@{}}
\toprule
\multicolumn{1}{c|}{\multirow{2}{*}{\textbf{Method}}} &
  \multicolumn{6}{c|}{\textbf{OLMo-3-7B}} &
  \multicolumn{6}{c}{\textbf{OLMo-3.1-32B}} \\ \cmidrule(l){2-13} 
\multicolumn{1}{c|}{} &
  \textbf{Person} &
  \textbf{Location} &
  \textbf{Organization} &
  \textbf{Art} &
  \multicolumn{1}{c|}{\textbf{Product}} &
  \multicolumn{1}{l|}{\textbf{Average}} &
  \textbf{Person} &
  \textbf{Location} &
  \textbf{Organization} &
  \textbf{Art} &
  \multicolumn{1}{c|}{\textbf{Product}} &
  \multicolumn{1}{l}{\textbf{Average}} \\ \midrule
\multicolumn{13}{c}{\textit{All Entities}} \\ \midrule
\multicolumn{1}{l|}{\textbf{Wikipedia}} &
  0.820 &
  \myhigh 0.826 &
  0.732 &
  \myhigh 0.713 &
  \multicolumn{1}{c|}{\myhigh 0.688} &
  \multicolumn{1}{c|}{{\ul 0.755}} &
  0.820 &
  0.826 &
  0.732 &
  0.713 &
  \multicolumn{1}{c|}{0.688} &
  0.755 \\
\multicolumn{1}{l|}{\textbf{Directly}} &
  0.729 &
  0.710 &
  0.593 &
  0.460 &
  \multicolumn{1}{c|}{0.580} &
  \multicolumn{1}{c|}{0.614} &
  0.758 &
  0.815 &
  0.761 &
  0.725 &
  \multicolumn{1}{c|}{0.708} &
  0.753 \\
\multicolumn{1}{l|}{\textbf{Comparison}} &
  \myhigh 0.823 &
  0.798 &
  \myhigh 0.748 &
  0.672 &
  \multicolumn{1}{c|}{0.645} &
  \multicolumn{1}{c|}{0.737} &
  \myhigh 0.837 &
  \myhigh 0.857 &
  \myhigh 0.784 &
  \myhigh 0.775 &
  \multicolumn{1}{c|}{\myhigh 0.722} &
  {\ul 0.795} \\ \midrule
\multicolumn{13}{c}{\textit{Sparse Entities}} \\ \midrule
\multicolumn{1}{l|}{\textbf{Wikipedia}} &
  0.003 &
  -0.022 &
  0.119 &
  0.137 &
  \multicolumn{1}{c|}{0.269} &
  \multicolumn{1}{c|}{0.101} &
  0.003 &
  -0.022 &
  0.119 &
  0.137 &
  \multicolumn{1}{c|}{0.269} &
  0.101 \\
\multicolumn{1}{l|}{\textbf{Directly}} &
  0.082 &
  \myhigh 0.229 &
  0.269 &
  \myhigh 0.225 &
  \multicolumn{1}{c|}{0.373} &
  \multicolumn{1}{c|}{0.235} &
  \myhigh 0.204 &
  0.113 &
  0.283 &
  0.145 &
  \multicolumn{1}{c|}{0.470} &
  0.243 \\
\multicolumn{1}{l|}{\textbf{Comparison}} &
  \myhigh 0.146 &
  0.226 &
  \myhigh 0.304 &
  0.201 &
  \multicolumn{1}{c|}{\myhigh 0.541} &
  \multicolumn{1}{c|}{{\ul 0.283}} &
  0.137 &
  \myhigh 0.241 &
  \myhigh 0.378 &
  \myhigh 0.307 &
  \multicolumn{1}{c|}{\myhigh 0.481} &
  {\ul 0.308} \\ \midrule
\multicolumn{13}{c}{\textit{Popular Entities}} \\ \midrule
\multicolumn{1}{l|}{\textbf{Wikipedia}} &
  \myhigh 0.557 &
  \myhigh 0.701 &
  0.280 &
  0.160 &
  \multicolumn{1}{c|}{0.343} &
  \multicolumn{1}{c|}{0.408} &
  0.557 &
  \myhigh 0.701 &
  0.280 &
  0.160 &
  \multicolumn{1}{c|}{0.343} &
  0.408 \\
\multicolumn{1}{l|}{\textbf{Directly}} &
  0.480 &
  0.381 &
  0.080 &
  0.287 &
  \multicolumn{1}{c|}{0.286} &
  \multicolumn{1}{c|}{0.302} &
  0.126 &
  0.564 &
  0.369 &
  0.317 &
  \multicolumn{1}{c|}{0.416} &
  0.358 \\
\multicolumn{1}{l|}{\textbf{Comparison}} &
  0.533 &
  0.655 &
  \myhigh 0.337 &
  \myhigh 0.336 &
  \multicolumn{1}{c|}{\myhigh 0.393} &
  \multicolumn{1}{c|}{{\ul 0.450}} &
  \myhigh 0.602 &
  0.696 &
  \myhigh 0.427 &
  \myhigh 0.389 &
  \multicolumn{1}{c|}{\myhigh 0.434} &
  {\ul 0.509} \\ \bottomrule
\end{tabular}%
}
\end{table*}

%% file: Main.bib
@article{lehmann2025knowing,
  title={Knowing the Facts but Choosing the Shortcut: Understanding How Large Language Models Compare Entities},
  author={Lehmann, Hans Hergen and Lee, Jae Hee and Schockaert, Steven and Wermter, Stefan},
  journal={arXiv preprint arXiv:2510.16815},
  year={2025}
}

@inproceedings{wang-etal-2023-causal,
    title = "A Causal View of Entity Bias in (Large) Language Models",
    author = "Wang, Fei  and
      Mo, Wenjie  and
      Wang, Yiwei  and
      Zhou, Wenxuan  and
      Chen, Muhao",
    editor = "Bouamor, Houda  and
      Pino, Juan  and
      Bali, Kalika",
    booktitle = "Findings of the Association for Computational Linguistics: EMNLP 2023",
    month = dec,
    year = "2023",
    address = "Singapore",
    publisher = "Association for Computational Linguistics",
    url = "https://aclanthology.org/2023.findings-emnlp.1013/",
    doi = "10.18653/v1/2023.findings-emnlp.1013",
    pages = "15173--15184"
}

@article{10.1145/3597307,
author = {Navigli, Roberto and Conia, Simone and Ross, Bj\"{o}rn},
title = {Biases in Large Language Models: Origins, Inventory, and Discussion},
year = {2023},
issue_date = {June 2023},
publisher = {Association for Computing Machinery},
address = {New York, NY, USA},
volume = {15},
number = {2},
issn = {1936-1955},
url = {https://doi.org/10.1145/3597307},
doi = {10.1145/3597307},
journal = {J. Data and Information Quality},
month = jun,
articleno = {10},
numpages = {21}
}

@inproceedings{prato-etal-2024-large,
    title = "Do Large Language Models Know How Much They Know?",
    author = "Prato, Gabriele  and
      Huang, Jerry  and
      Parthasarathi, Prasanna  and
      Sodhani, Shagun  and
      Chandar, Sarath",
    editor = "Al-Onaizan, Yaser  and
      Bansal, Mohit  and
      Chen, Yun-Nung",
    booktitle = "Proceedings of the 2024 Conference on Empirical Methods in Natural Language Processing",
    month = nov,
    year = "2024",
    address = "Miami, Florida, USA",
    publisher = "Association for Computational Linguistics",
    url = "https://aclanthology.org/2024.emnlp-main.348/",
    doi = "10.18653/v1/2024.emnlp-main.348",
    pages = "6054--6070"
}

@inproceedings{dodge-etal-2021-documenting,
    title = "Documenting Large Webtext Corpora: A Case Study on the Colossal Clean Crawled Corpus",
    author = "Dodge, Jesse  and
      Sap, Maarten  and
      Marasovi{\'c}, Ana  and
      Agnew, William  and
      Ilharco, Gabriel  and
      Groeneveld, Dirk  and
      Mitchell, Margaret  and
      Gardner, Matt",
    editor = "Moens, Marie-Francine  and
      Huang, Xuanjing  and
      Specia, Lucia  and
      Yih, Scott Wen-tau",
    booktitle = "Proceedings of the 2021 Conference on Empirical Methods in Natural Language Processing",
    month = nov,
    year = "2021",
    address = "Online and Punta Cana, Dominican Republic",
    publisher = "Association for Computational Linguistics",
    url = "https://aclanthology.org/2021.emnlp-main.98/",
    doi = "10.18653/v1/2021.emnlp-main.98",
    pages = "1286--1305"
}

@article{kadavath2022language,
  title={Language models (mostly) know what they know},
  author={Kadavath, Saurav and Conerly, Tom and Askell, Amanda and Henighan, Tom and Drain, Dawn and Perez, Ethan and Schiefer, Nicholas and Hatfield-Dodds, Zac and DasSarma, Nova and Tran-Johnson, Eli and others},
  journal={arXiv preprint arXiv:2207.05221},
  year={2022}
}

@article{olmo2025olmo,
  title={Olmo 3},
  author={Olmo, Team and Ettinger, Allyson and Bertsch, Amanda and Kuehl, Bailey and Graham, David and Heineman, David and Groeneveld, Dirk and Brahman, Faeze and Timbers, Finbarr and Ivison, Hamish and others},
  journal={arXiv preprint arXiv:2512.13961},
  year={2025}
}

@inproceedings{liuinfini,
  title={Infini-gram: Scaling Unbounded n-gram Language Models to a Trillion Tokens},
  author={Liu, Jiacheng and Min, Sewon and Zettlemoyer, Luke and Choi, Yejin and Hajishirzi, Hannaneh},
  booktitle={First Conference on Language Modeling}
}

@article{achiam2023gpt,
  title={Gpt-4 technical report},
  author={Achiam, Josh and Adler, Steven and Agarwal, Sandhini and Ahmad, Lama and Akkaya, Ilge and Aleman, Florencia Leoni and Almeida, Diogo and Altenschmidt, Janko and Altman, Sam and Anadkat, Shyamal and others},
  journal={arXiv preprint arXiv:2303.08774},
  year={2023}
}

@inproceedings{mallen-etal-2023-trust,
    title = "When Not to Trust Language Models: Investigating Effectiveness of Parametric and Non-Parametric Memories",
    author = "Mallen, Alex  and
      Asai, Akari  and
      Zhong, Victor  and
      Das, Rajarshi  and
      Khashabi, Daniel  and
      Hajishirzi, Hannaneh",
    editor = "Rogers, Anna  and
      Boyd-Graber, Jordan  and
      Okazaki, Naoaki",
    booktitle = "Proceedings of the 61st Annual Meeting of the Association for Computational Linguistics (Volume 1: Long Papers)",
    month = jul,
    year = "2023",
    address = "Toronto, Canada",
    publisher = "Association for Computational Linguistics",
    url = "https://aclanthology.org/2023.acl-long.546/",
    doi = "10.18653/v1/2023.acl-long.546",
    pages = "9802--9822"
}

@inbook{pytorch,
author = {Paszke, Adam and Gross, Sam and Massa, Francisco and Lerer, Adam and Bradbury, James and Chanan, Gregory and Killeen, Trevor and Lin, Zeming and Gimelshein, Natalia and Antiga, Luca and Desmaison, Alban and K\"{o}pf, Andreas and Yang, Edward and DeVito, Zach and Raison, Martin and Tejani, Alykhan and Chilamkurthy, Sasank and Steiner, Benoit and Fang, Lu and Bai, Junjie and Chintala, Soumith},
title = {PyTorch: an imperative style, high-performance deep learning library},
year = {2019},
publisher = {Curran Associates Inc.},
address = {Red Hook, NY, USA},
booktitle = {Proceedings of the 33rd International Conference on Neural Information Processing Systems},
articleno = {721},
numpages = {12}
}

@inproceedings{wolf-etal-2020-transformers,
    title = "Transformers: State-of-the-Art Natural Language Processing",
    author = "Wolf, Thomas  and
      Debut, Lysandre  and
      Sanh, Victor  and
      Chaumond, Julien  and
      Delangue, Clement  and
      Moi, Anthony  and
      Cistac, Pierric  and
      Rault, Tim  and
      Louf, Remi  and
      Funtowicz, Morgan  and
      Davison, Joe  and
      Shleifer, Sam  and
      von Platen, Patrick  and
      Ma, Clara  and
      Jernite, Yacine  and
      Plu, Julien  and
      Xu, Canwen  and
      Le Scao, Teven  and
      Gugger, Sylvain  and
      Drame, Mariama  and
      Lhoest, Quentin  and
      Rush, Alexander",
    editor = "Liu, Qun  and
      Schlangen, David",
    booktitle = "Proceedings of the 2020 Conference on Empirical Methods in Natural Language Processing: System Demonstrations",
    month = oct,
    year = "2020",
    address = "Online",
    publisher = "Association for Computational Linguistics",
    url = "https://aclanthology.org/2020.emnlp-demos.6/",
    doi = "10.18653/v1/2020.emnlp-demos.6",
    pages = "38--45"
}

@inproceedings{wangself,
  title={Self-Consistency Improves Chain of Thought Reasoning in Language Models},
  author={Wang, Xuezhi and Wei, Jason and Schuurmans, Dale and Le, Quoc V and Chi, Ed H and Narang, Sharan and Chowdhery, Aakanksha and Zhou, Denny},
  booktitle={The Eleventh International Conference on Learning Representations}
}

@article{bradley-terry,
 ISSN = {00063444, 14643510},
 URL = {http://www.jstor.org/stable/2334029},
 author = {Ralph Allan Bradley and Milton E. Terry},
 journal = {Biometrika},
 number = {3/4},
 pages = {324--345},
 publisher = {[Oxford University Press, Biometrika Trust]},
 title = {Rank Analysis of Incomplete Block Designs: I. The Method of Paired Comparisons},
 urldate = {2026-02-04},
 volume = {39},
 year = {1952}
}

@article{grattafiori2024llama,
  title={The llama 3 herd of models},
  author={Grattafiori, Aaron and Dubey, Abhimanyu and Jauhri, Abhinav and Pandey, Abhinav and Kadian, Abhishek and Al-Dahle, Ahmad and Letman, Aiesha and Mathur, Akhil and Schelten, Alan and Vaughan, Alex and others},
  journal={arXiv preprint arXiv:2407.21783},
  year={2024}
}

@article{yang2025qwen3,
  title={Qwen3 technical report},
  author={Yang, An and Li, Anfeng and Yang, Baosong and Zhang, Beichen and Hui, Binyuan and Zheng, Bo and Yu, Bowen and Gao, Chang and Huang, Chengen and Lv, Chenxu and others},
  journal={arXiv preprint arXiv:2505.09388},
  year={2025}
}

@article{team2024gemma,
  title={Gemma: Open models based on gemini research and technology},
  author={Team, Gemma and Mesnard, Thomas and Hardin, Cassidy and Dadashi, Robert and Bhupatiraju, Surya and Pathak, Shreya and Sifre, Laurent and Rivi{\`e}re, Morgane and Kale, Mihir Sanjay and Love, Juliette and others},
  journal={arXiv preprint arXiv:2403.08295},
  year={2024}
}

@inproceedings{huang-etal-2023-large,
    title = "Large Language Models Can Self-Improve",
    author = "Huang, Jiaxin  and
      Gu, Shixiang  and
      Hou, Le  and
      Wu, Yuexin  and
      Wang, Xuezhi  and
      Yu, Hongkun  and
      Han, Jiawei",
    editor = "Bouamor, Houda  and
      Pino, Juan  and
      Bali, Kalika",
    booktitle = "Proceedings of the 2023 Conference on Empirical Methods in Natural Language Processing",
    month = dec,
    year = "2023",
    address = "Singapore",
    publisher = "Association for Computational Linguistics",
    url = "https://aclanthology.org/2023.emnlp-main.67/",
    doi = "10.18653/v1/2023.emnlp-main.67",
    pages = "1051--1068"
}

@inproceedings{soldaini-etal-2024-dolma,
    title = "Dolma: an Open Corpus of Three Trillion Tokens for Language Model Pretraining Research",
    author = "Soldaini, Luca  and
      Kinney, Rodney  and
      Bhagia, Akshita  and
      Schwenk, Dustin  and
      Atkinson, David  and
      Authur, Russell  and
      Bogin, Ben  and
      Chandu, Khyathi  and
      Dumas, Jennifer  and
      Elazar, Yanai  and
      Hofmann, Valentin  and
      Jha, Ananya  and
      Kumar, Sachin  and
      Lucy, Li  and
      Lyu, Xinxi  and
      Lambert, Nathan  and
      Magnusson, Ian  and
      Morrison, Jacob  and
      Muennighoff, Niklas  and
      Naik, Aakanksha  and
      Nam, Crystal  and
      Peters, Matthew  and
      Ravichander, Abhilasha  and
      Richardson, Kyle  and
      Shen, Zejiang  and
      Strubell, Emma  and
      Subramani, Nishant  and
      Tafjord, Oyvind  and
      Walsh, Evan  and
      Zettlemoyer, Luke  and
      Smith, Noah  and
      Hajishirzi, Hannaneh  and
      Beltagy, Iz  and
      Groeneveld, Dirk  and
      Dodge, Jesse  and
      Lo, Kyle",
    editor = "Ku, Lun-Wei  and
      Martins, Andre  and
      Srikumar, Vivek",
    booktitle = "Proceedings of the 62nd Annual Meeting of the Association for Computational Linguistics (Volume 1: Long Papers)",
    month = aug,
    year = "2024",
    address = "Bangkok, Thailand",
    publisher = "Association for Computational Linguistics",
    url = "https://aclanthology.org/2024.acl-long.840/",
    doi = "10.18653/v1/2024.acl-long.840",
    pages = "15725--15788"
}

@article{spearman-correlation,
	title        = {The Proof and Measurement of Association between Two Things},
	author       = {C. Spearman},
	year         = 1904,
	journal      = {The American Journal of Psychology},
	publisher    = {University of Illinois Press},
	volume       = 15,
	number       = 1,
	pages        = {72--101},
	issn         = {00029556},
	url          = {http://www.jstor.org/stable/1412159},
	urldate      = {2025-02-06}
}

@article{shi2023detecting,
  title={Detecting pretraining data from large language models},
  author={Shi, Weijia and Ajith, Anirudh and Xia, Mengzhou and Huang, Yangsibo and Liu, Daogao and Blevins, Terra and Chen, Danqi and Zettlemoyer, Luke},
  journal={arXiv preprint arXiv:2310.16789},
  year={2023}
}

@inproceedings{zhang-etal-2024-pretraining,
    title = "Pretraining Data Detection for Large Language Models: A Divergence-based Calibration Method",
    author = "Zhang, Weichao  and
      Zhang, Ruqing  and
      Guo, Jiafeng  and
      de Rijke, Maarten  and
      Fan, Yixing  and
      Cheng, Xueqi",
    editor = "Al-Onaizan, Yaser  and
      Bansal, Mohit  and
      Chen, Yun-Nung",
    booktitle = "Proceedings of the 2024 Conference on Empirical Methods in Natural Language Processing",
    month = nov,
    year = "2024",
    address = "Miami, Florida, USA",
    publisher = "Association for Computational Linguistics",
    url = "https://aclanthology.org/2024.emnlp-main.300/",
    doi = "10.18653/v1/2024.emnlp-main.300",
    pages = "5263--5274"
}

@inproceedings{balloccu-etal-2024-leak,
    title = "Leak, Cheat, Repeat: Data Contamination and Evaluation Malpractices in Closed-Source {LLM}s",
    author = "Balloccu, Simone  and
      Schmidtov{\'a}, Patr{\'i}cia  and
      Lango, Mateusz  and
      Dusek, Ondrej",
    editor = "Graham, Yvette  and
      Purver, Matthew",
    booktitle = "Proceedings of the 18th Conference of the European Chapter of the Association for Computational Linguistics (Volume 1: Long Papers)",
    month = mar,
    year = "2024",
    address = "St. Julian{'}s, Malta",
    publisher = "Association for Computational Linguistics",
    url = "https://aclanthology.org/2024.eacl-long.5/",
    doi = "10.18653/v1/2024.eacl-long.5",
    pages = "67--93"
}

@inproceedings{shuster-etal-2022-language,
    title = "Language Models that Seek for Knowledge: Modular Search {\&} Generation for Dialogue and Prompt Completion",
    author = "Shuster, Kurt  and
      Komeili, Mojtaba  and
      Adolphs, Leonard  and
      Roller, Stephen  and
      Szlam, Arthur  and
      Weston, Jason",
    editor = "Goldberg, Yoav  and
      Kozareva, Zornitsa  and
      Zhang, Yue",
    booktitle = "Findings of the Association for Computational Linguistics: EMNLP 2022",
    month = dec,
    year = "2022",
    address = "Abu Dhabi, United Arab Emirates",
    publisher = "Association for Computational Linguistics",
    url = "https://aclanthology.org/2022.findings-emnlp.27/",
    doi = "10.18653/v1/2022.findings-emnlp.27",
    pages = "373--393"
}

@inproceedings{carlini2021extracting,
  title={Extracting training data from large language models},
  author={Carlini, Nicholas and Tramer, Florian and Wallace, Eric and Jagielski, Matthew and Herbert-Voss, Ariel and Lee, Katherine and Roberts, Adam and Brown, Tom and Song, Dawn and Erlingsson, Ulfar and others},
  booktitle={30th USENIX security symposium (USENIX Security 21)},
  pages={2633--2650},
  year={2021}
}

@inproceedings{10.1145/3357713.3384290,
author = {Feldman, Vitaly},
title = {Does learning require memorization? a short tale about a long tail},
year = {2020},
isbn = {9781450369794},
publisher = {Association for Computing Machinery},
address = {New York, NY, USA},
url = {https://doi.org/10.1145/3357713.3384290},
doi = {10.1145/3357713.3384290},
booktitle = {Proceedings of the 52nd Annual ACM SIGACT Symposium on Theory of Computing},
pages = {954–959},
numpages = {6},
location = {Chicago, IL, USA},
series = {STOC 2020}
}

@inproceedings{petroni-etal-2019-language,
    title = "Language Models as Knowledge Bases?",
    author = {Petroni, Fabio  and
      Rockt{\"a}schel, Tim  and
      Riedel, Sebastian  and
      Lewis, Patrick  and
      Bakhtin, Anton  and
      Wu, Yuxiang  and
      Miller, Alexander},
    editor = "Inui, Kentaro  and
      Jiang, Jing  and
      Ng, Vincent  and
      Wan, Xiaojun",
    booktitle = "Proceedings of the 2019 Conference on Empirical Methods in Natural Language Processing and the 9th International Joint Conference on Natural Language Processing (EMNLP-IJCNLP)",
    month = nov,
    year = "2019",
    address = "Hong Kong, China",
    publisher = "Association for Computational Linguistics",
    url = "https://aclanthology.org/D19-1250/",
    doi = "10.18653/v1/D19-1250",
    pages = "2463--2473"
}

@inproceedings{tong2025pde_survey,
author = {Tong, Ziyi and Sun, Feifei and Nguyen, Le Minh},
title = {Pretraining Data Exposure in Large Language Models: A Survey of Membership Inference, Data Contamination, and Security Implications},
year = {2025},
isbn = {978-3-031-97143-3},
publisher = {Springer-Verlag},
address = {Berlin, Heidelberg},
url = {https://doi.org/10.1007/978-3-031-97144-0_14},
doi = {10.1007/978-3-031-97144-0_14},
booktitle = {Natural Language Processing and Information Systems: 30th International Conference on Applications of Natural Language to Information Systems, NLDB 2025, Kanazawa, Japan, July 4–6, 2025, Proceedings, Part II},
pages = {152–162},
numpages = {11},
location = {Kanazawa, Japan}
}

@article{pozzi2025distillation_exposure_bias,
author = {Pozzi, Andrea and Incremona, Alessandro and Tessera, Daniele and Toti, Daniele},
title = {Mitigating exposure bias in large language model distillation: an imitation learning approach},
year = {2025},
issue_date = {Jun 2025},
publisher = {Springer-Verlag},
address = {Berlin, Heidelberg},
volume = {37},
number = {18},
issn = {0941-0643},
url = {https://doi.org/10.1007/s00521-025-11162-0},
doi = {10.1007/s00521-025-11162-0},
journal = {Neural Comput. Appl.},
month = mar,
pages = {12013–12029},
numpages = {17}
}

@inproceedings{yu2023self_information_update,
    title = "Information Association for Language Model Updating by Mitigating {LM}-Logical Discrepancy",
    author = "Yu, Pengfei  and
      Ji, Heng",
    editor = "Barak, Libby  and
      Alikhani, Malihe",
    booktitle = "Proceedings of the 28th Conference on Computational Natural Language Learning",
    month = nov,
    year = "2024",
    address = "Miami, FL, USA",
    publisher = "Association for Computational Linguistics",
    url = "https://aclanthology.org/2024.conll-1.10/",
    doi = "10.18653/v1/2024.conll-1.10",
    pages = "117--129"
}
